\documentclass{llncs}

\usepackage{latexsym}
\usepackage[table, dvipsnames]{xcolor}
\usepackage{booktabs}
\usepackage{graphicx}
\usepackage{multirow}
\usepackage{amsfonts}
\usepackage{tabularx}
\usepackage{array}
\usepackage{tikz}
\usepackage{multirow}
\usepackage{amsmath}
\usepackage{floatrow}

\newcolumntype{C}[1]{>{\centering\arraybackslash}p{#1}}
\newcolumntype{W}{>{\centering\arraybackslash}p{1.2cm}}

\title{Aspect-Based Sentiment Analysis Using a Two-Step Neural Network Architecture}
\author{
  Soufian Jebbara and Philipp Cimiano
  }
\institute{  
  Semantic Computing Group \\ Cognitive Interaction Technology -- Center of Excellence (CITEC) \\ Bielefeld University, Germany
}

\begin{document}

\maketitle
\begin{abstract}
The World Wide Web holds a wealth of information in the form of unstructured texts such as customer reviews for products, events and more.
By extracting and analyzing the expressed opinions in customer reviews in a fine-grained way, valuable opportunities and insights for customers and businesses can be gained.

We propose a neural network based system to address the task of Aspect-Based Sentiment Analysis to compete in Task 2 of the ESWC-2016 Challenge on Semantic Sentiment Analysis.
Our proposed architecture divides the task in two subtasks: aspect term extraction and aspect-specific sentiment extraction.
This approach is flexible in that it allows to address each subtask independently.
As a first step, a recurrent neural network is used to extract aspects from a text by framing the problem as a sequence labeling task.
In a second step, a recurrent network processes each extracted aspect with respect to its context and predicts a sentiment label.
The system uses pretrained semantic word embedding features which we experimentally enhance with semantic knowledge extracted from WordNet.
Further features extracted from SenticNet prove to be beneficial for the extraction of sentiment labels.
As the best performing system in its category, our proposed system proves to be an effective approach for the Aspect-Based Sentiment Analysis.
\end{abstract}

\section{Introduction}
The World Wide Web contains customer reviews for all kinds of topics and entities such as products, movies, events, restaurants and more.
The wealth of information that is expressed in these reviews in the form of the writer's opinion offers valuable opportunities and insights for customers and businesses altogether.
However, due to the vast amounts of customer reviews that are available in the Web, the manual extraction and analysis of these opinions is infeasible and thus requires automated tools.
First attempts to extract opinions automatically have focused on extracting an overall polarity on a document or sentence level.
This, however, is a too coarse-grained approach as it neglects huge amounts of information in these reviews.

In a more fine-grained way, Sentiment analysis can be regarded as a relation extraction problem in which the sentiment of some opinion holder towards a certain aspect of a product needs to be extracted.
The following example clearly shows that the mere extraction of an overall polarity for a sentence is not sufficient:
\\\\
\tikzstyle{every picture}+=[remember picture]
\tikzstyle{none} = [shape=rectangle,inner sep=2pt,outer sep=1pt,text depth=0pt]
\tikzstyle{sentiment} = [shape=rectangle,inner sep=2pt,outer sep=1pt,text depth=7pt]
\tikzstyle{aspect} = [draw,shape=rectangle,inner sep=2pt,outer sep=1pt,text depth=0pt]
\tikzstyle{opinion} = [draw,dashed,line width=0.8pt,shape=rectangle,inner sep=2pt,outer sep=1pt,text depth=0pt]
\tikz\node[none]{The}; \tikz\node[aspect](a1){serrated portion}; \tikz\node[none]{of the}; \tikz\node[aspect](a2){blade}; \tikz\node[none]{is}; \tikz\node[opinion](s1){sharp};\tikz\node[sentiment](sent1){\textit{pos}}; \tikz\node[none]{but the}; \tikz\node[aspect](a3){straight edge}; \tikz\node[none]{is}; \tikz\node[opinion](s2){marginal at best};\tikz\node[sentiment](sent2){\textit{neg}};\tikz\node[none]{.};
\begin{tikzpicture}[overlay]
  \path[->,black,semithick](s1) edge [out=160, in=20] (a1);
  \path[->,black,semithick](s2) edge [out=350, in=200] (a3);
\end{tikzpicture}
\\\\
where aspect terms are outlined with solid boxes, opinion phrases with dashed ones, opinion polarities are displayed as superscripts, and aspect-opinion dependencies are depicted as arrows.
Sentiment analysis needs to be regarded thus on a more fine-grained level that allows to assign sentiments to individual aspects in order to extract complex opinions more accurately.

In this work, we present a system that competes in the ESWC 2016 Challenge on Semantic Sentiment Analysis addressing the task of Aspect-Based Sentiment Analysis.
The goal of this task is to extract a set of aspect terms with their respective binary polarities (\texttt{positive} and \texttt{negative}) from a given sentence.
The sentences in the overall dataset are extracted from online reviews from different domains (restaurants, laptops and hotels).
We approach the problem in two steps: i) the extraction of aspect terms and ii) the assignment of a polarity label to each extracted aspect term.
Following this approach, we design a modular, neural network based architecture that is easy to extend.

In the following, we give a brief overview of related work in the field of aspect-based sentiment analysis.
Afterwards, we present our overall system and describe its two main components and the features we employ.
We further analyse the performance of our architecture on both subtasks and give insights into its predictive performance.
Lastly, we conclude the paper and give suggestions for further improvements.

\section{Related Work}

Our work is inspired by different related approaches for sentiment analysis.
Overall, our work is in line with the growing interest of providing more fine-grained, aspect-based sentiment analysis \cite{lakkaraju2014aspect,Klinger2013b,Pontiki2015}, going beyond a mere text classification or regression problem that aims at predicting an overall sentiment for a text.  

San Vicente et al. \cite{agerri2015elixa} present a system that addresses opinion target extraction as a sequence labeling problem based on a perceptron algorithm with local features.
The extraction of a sentiment polarity for an extracted opinion target is performed using an SVM. The approach uses a window of words around a given opinion target and classifies it based on a set of features such as word clusters, Part-of-Speech tags and polarity lexicon features.

Toh and Wang \cite{Toh2014} propose a Conditional Random Field (CRF) as a sequence labeler that includes a variety of features such as POS tags and dependencies, word clusters and WordNet taxonomies.
Additionally, the authors employ a logistic regression classifier to address aspect term polarity classification.

Jakob and Gurevych \cite{Jakob2010} follow a very similar approach that addresses opinion target extraction as a sequence labeling problem using CRFs.
Their approach includes features derived from words, POS tags and dependency paths, and performs well in a single and cross-domain setting.

Klinger and Cimiano \cite{Klinger2013a,Klinger2013b} have modeled the task of joint aspect and opinion term extraction using probabilistic graphical models and rely on Markov Chain Monte Carlo methods for inference.
They have demonstrated the impact of a joint architecture on the task with a strong impact on the extraction of aspect terms, but less so for the extraction of opinion terms.

Lakkaraju et al. \cite{lakkaraju2014aspect} present a recursive neural network architecture that is capable of extracting multiple aspect categories\footnote{Here, we distinguish between the terminologies of aspect \emph{category} extraction and aspect \emph{term} extraction:
The set of possible aspect categories is predefined and rather small (e.g. \texttt{Price, Battery, Accessories, Display, Portability, Camera}), while aspect terms can take many shapes (e.g. ``\textit{sake menu}'', ``\textit{wine selection}'' or ``\textit{French Onion soup}'').} and their respective sentiments jointly in one model or separately using two softmax classifiers.
They show that the joint modeling of aspect categories and sentiments is beneficial for the predictive performance of their system.

Another way to address opinion extraction is the summarization of reviews.
Hu and Liu \cite{Hu2004} present an approach that summarizes reviews based on the product features for which an opinion is expressed using data mining and natural language processing techniques.
Similarly, Titov and McDonald \cite{Titov2008} describe a statistical model for joint aspect and sentiment modeling for the summarization of reviews. The method is based on Multi-Grain Latent Dirichlet Allocation which models global and local topics extended by a Multi-Aspect Sentiment Model.

Lastly, the general idea expressed in this paper to incorporate semantic web technologies in a machine learning framework for sentiment analysis is rooted in previous contributions of ESWC Challenges \cite{SchoutenF15,AprosioCDR15,DragoniTP14,ChungWT14}.

\section{Aspect-Based Sentiment Analysis}
We follow a two-step approach in designing a system that is capable of extracting a writer's sentiment towards certain aspects of an entity (such as a product or restaurant).
As a first step, given a text, the system extracts explicitly expressed aspects\footnote{Parts of a sentence that refer to an aspect of the product, event, entity, etc.} in this text.
Secondly, each extracted aspect term is processed individually and a sentiment value is assigned given the context of the aspect term.

This two-step approach allows us to extract an arbitrary amount of aspects from a text.
Additionally, by decoupling the aspect extraction from the sentiment extraction, the system is also applicable to settings where aspect terms are already given and only the individual sentiments towards these aspects need to be extracted.
The following sections elaborate on our design and feature choices for our aspect and sentiment extraction components.

\subsection{Features}
\label{sec:features}
In this section, we describe the features we use to address aspect term extraction and aspect-specific sentiment extraction.
For both sub tasks, we lowercase each input sentence and tokenize it using a simple regular expression in a pre-processing step.
We do not remove punctuations or stopwords, but keep them intact.

\subsubsection{Word Embeddings}
The most important features that we use are pretrained word embeddings which have been successfully employed in numerous NLP tasks \cite{collobert2011natural,santos2014learning,Le2014,Mikolov2013Efficient,pennington2014glove}.
We use the skip-gram model \cite{Mikolov2013Efficient} with negative sampling on a huge corpus of $\approx 83$ million Amazon reviews \cite{McAPanLes15,McATarShiHen15} to compute 100 dimensional word embeddings.
In total, our computation of word embeddings yields vectors for $\approx$ 1 million words.
For this work, however, we reduce this vocabulary to only contain the 100,000 most frequent words.
The resulting vocabulary is denoted as $V$.

In a pre-processing step, we replace rare words that appear less than 10 times in our dataset with a special token \texttt{<UNK>} and learn a \emph{placeholder} vector for this token.
At test time, we use this token as a replacement for Out-of-Vocabulary words.
The sequence of word embedding vectors for a sentence with words\footnote{For a more convenient notation, we use words and their respective indices interchangeably.} $1 \ldots N$ is denoted as:
$$[w]_1^N= \{w_1, \ldots, w_N\} \text{ with } w_i \in \mathbb{R}^{100}.$$

By using this domain-specific dataset we expect to obtain embeddings that capture the semantics of each word for our targeted domain more closely than embeddings trained on domain-independent data.
A welcomed side effect of using this huge dataset of reviews is that we also obtain word embeddings for misspelled forms of a word that appear commonly in reviews.
As shown in Table \ref{tab:misspelled}, the learned representation of a misspelled word is in many cases very close\footnote{We use the euclidean vector distance as a distance measure.} to its correctly spelled counterpart.
\begin{table}[t]
\centering
\begin{tabular}{lp{0.5cm}C{1.5cm}C{1.5cm}C{1.5cm}}
\toprule
\multicolumn{1}{l}{Word}           & & speed  & quality  & display  \\
\midrule
\multirow{3}{*}{Nearest Neighbors} && \textit{spped}  & \textit{qualtiy}  & displays \\
				  && speeds & \textit{qualilty} & \textit{diplay}   \\
				  && \textit{speeed} & \textit{qulaity}  & \textit{dislay}\\
\bottomrule
\end{tabular}
\vspace{0.5cm}
\caption{Three commonly used words in product reviews and their 3 nearest neighbors in the embedding space. Often, misspelled versions (\textit{italic}) of the original word are among its closest neighbors.}
\label{tab:misspelled}
\end{table}

Although our approach technically works without any features apart from word embeddings, we are interested in improving its performance by means of semantic web technology.
For that, we employ features derived from two graph-based semantic resources: WordNet and SenticNet.

\subsubsection{Retrofitting Word Embeddings to WordNet}
Although word embeddings have been shown to encode semantic and syntactic features of their respective words well \cite{Mikolov2013Distributed,santos2014learning,Mikolov2013Efficient}, we try to enhance their encoded semantics by using a lexical resource.
For this, we employ a technique called \emph{retrofitting} \cite{Faruqui2015}.
The idea behind retrofitting is to iteratively adapt precomputed word vectors to better fit the (lexical) relations modeled in a given lexical resource.
The graph-based algorithm gradually ``moves'' each word vector towards the word vectors of its neighboring nodes while still staying close to its original position.

Formally, following the notation by Faruqui et al. \cite{Faruqui2015}, let $V=\{v_1,\dots,v_{N_V}\}$ be the considered vocabulary of size $N_V$ and $\hat W=(\hat w_1,\dots,\hat w_{|V|})$ with $w_i \in \mathbb{R}^d$ are their respective precomputed word vectors.
$G=(V,E)$ is the graph of semantic relationships to which we want to fit the word vectors with $(v_i,v_j)\in E \subseteq V \times V$ denoting the edges between words.
With $ W=( w_1,\dots, w_{|V|})$ being the fitted word vectors, the algorithm tries to minimize the following objective function:
\begin{equation}
\Psi(\hat W, W)=\sum^{|V|}_{i=1}  \Bigg[\alpha_i||w_i-\hat w_i||^2 + \sum\limits_{(i,j) \in E} \beta_{ij}||w_i-w_j||^2\Bigg]
\end{equation}
The online update rule for each $w_i$ is then:
\begin{equation}
w_i=\frac{\sum_{j:(i,j)\in E} \beta_{ij}w_j + \alpha_i \hat w_i }{\sum_{j:(i,j)\in E} \beta_{ij} + \alpha_i  }
\end{equation}
where $\alpha$ and $\beta$ are parameters of the retrofitting procedure.

In this work, we chose WordNet \cite{fellbaum2005} as our lexico-semantic resource.
We construct a subgraph of the WordNet relations that links each word in our vocabulary to all its synonyms (lemma names) in the WordNet graph.
We set all $\alpha_i=1$ and all $\beta_{ij}=1/degree(i)$ and run the retrofitting algorithm for 10 iterations.
The resulting embeddings are still very similar to their original embeddings, yet incorporate part of the semantics of WordNet.
We investigate the benefit of using these retrofitted word embeddings in comparison to their original counterparts in section \ref{sec:experiments}.

\subsubsection{SenticNet}
\label{sec:sentic}
SenticNet 3 \cite{Cambria2014sentic} is a graph-based, concept-level resource for semantic and affective information.
For each of the 30,000 concepts that are part of the knowledge graph, SenticNet 3 provides real-valued scores for 5 \emph{sentics}: \texttt{pleasantness}, \texttt{attention}, \texttt{sensitivity}, \texttt{aptitude}, \texttt{polarity}.

We experimentally include the provided scores in our system as an additional input source that our networks can draw information from.
Since these sentics encode information about the semantics and polarity of a concept, the aspect-specific sentiment extraction component is expected to benefit from the additional information in particular.
For that, we construct a 5-dimensional feature vector $s_c$ for each concept $c$ that is represented in SenticNet 3.
We refer to these vectors as \emph{sentic vectors}.

Unfortunately, our system is not designed to process text on a concept level but only on a word level.
Therefore, we omit all multi-word concepts (e.g. \texttt{notice\_problem} or \texttt{beautiful\_music}) in SenticNet 3 and only keep single-word concepts (e.g. \texttt{experience} or \texttt{improvement}) that are part of our vocabulary $V$.
Doing that, we can treat the sentic vector $s_i$ as an additional word vector for the word $i$.
To account for Out-of-Vocabulary words during test time, we provide a default vector $s_{unk}=\mathbf{0}$. 
The sequence of sentic vectors for a sentence with words $1 \ldots N$ is denoted as:
$$[s]_1^N= \{s_1, \ldots, s_N\} \text{ with } s_i \in \mathbb{R}^{5}.$$

\subsubsection{Part-of-Speech Tags}
Apart from these word embeddings and sentic vectors, our system can incorporate other features as well.
For each word in a text, Part-of-Speech (POS) tags can be provided that might aid both the aspect extraction and aspect-specific sentiment extraction components.
When including POS tags, we employ a 1-of-K coding scheme that transforms each tag into a K-dimensional vector that represents this specific tag.
Specifically, we use the Stanford POS Tagger \cite{Manning2014} with a tag set of 45 tags.
These vectors are then concatenated with their respective word vectors before being fed to the extraction components.
The sequence of POS tag vectors for a sentence with words $1 \ldots N$ is denoted as:
$$[p]_1^N= \{p_1, \ldots, p_N\} \text{ with } p_i \in \mathbb{R}^{45}.$$

\subsection{Aspect Term Extraction}
\label{sec:model}
Our first step in extracting aspect-based sentiment from a text is the extraction of mentioned aspect terms.
We propose a system to extract an arbitrary number of aspect terms from a given text by framing the extraction as a sequence labeling problem.
For this, we encode expressed aspect terms using the IOB2 tagging scheme \cite{tksveenstra99eacl}.
According to this scheme, each word in our text receives one of 3 tags, namely \textbf{I}, \textbf{O} or \textbf{B} that indicate if the word is at the \textbf{B}eginning, \textbf{I}nside or \textbf{O}utside of an annotation:
\begin{center}
\small
\begin{tabular}{WWWWWWWW}
The & \textbf{sake} & \textbf{menu} & should & not & be & overlooked & !\\
O & \textbf{B} & \textbf{I}& O & O & O & O & O
\end{tabular}
\end{center}
This tagging scheme allows us to encode multiple non-overlapping aspect terms at once.
Ultimately, each tag is represented as a 1-of-K vector:
$$
I=\begin{bmatrix}
1\\
0\\
0
\end{bmatrix},
O=\begin{bmatrix}
0\\
1\\
0
\end{bmatrix},
B=\begin{bmatrix}
0\\
0\\
1
\end{bmatrix}.
$$

\begin{figure}
  \centering
  \includegraphics[width=0.75\columnwidth]{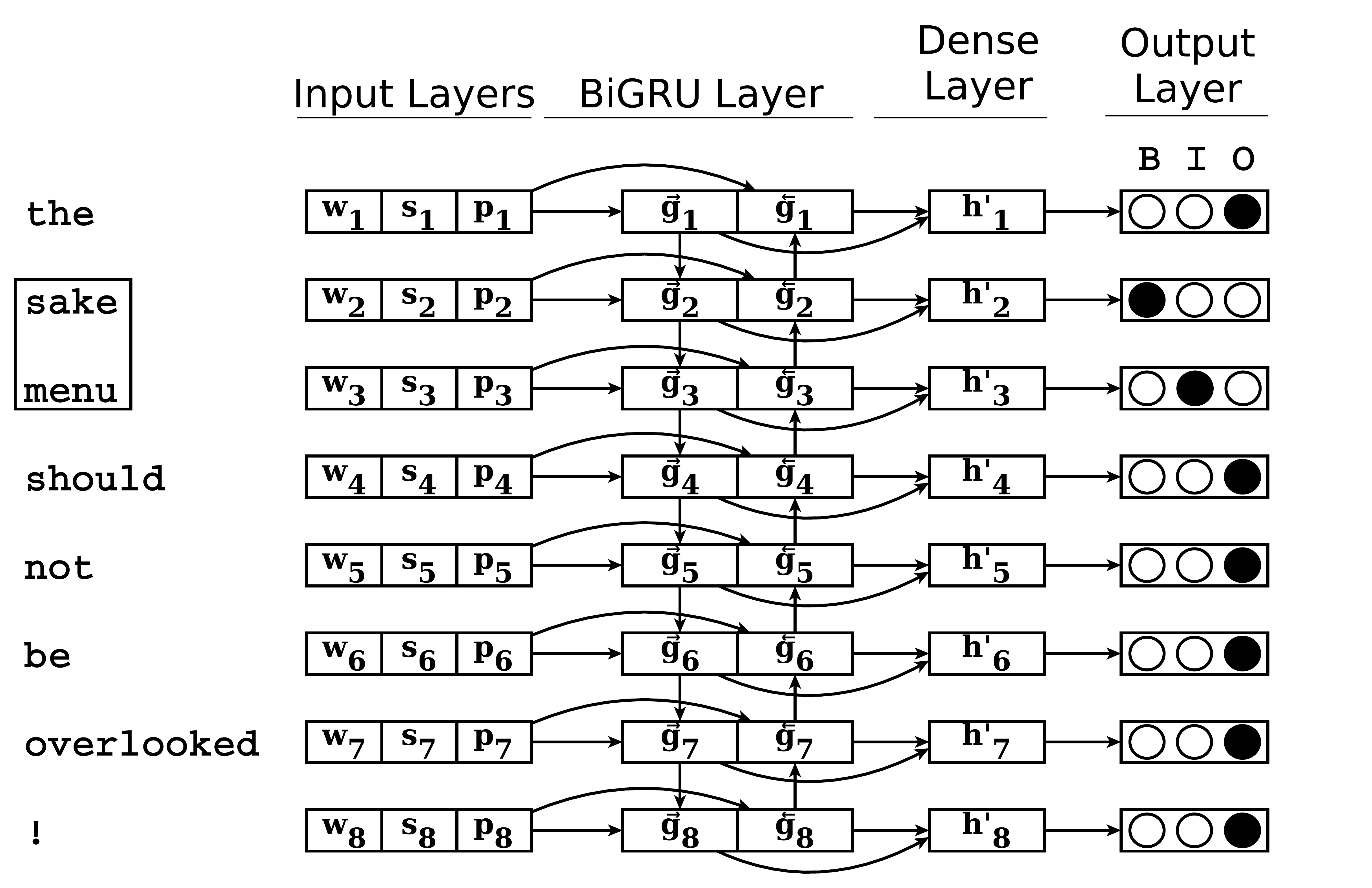}
  \caption{The aspect term extraction component. The network processes the input sentence as a sequence of word vectors $w_i$, sentic vectors $s_i$ and POS tags $p_i$ using a bidirectional GRU layer and regular feed-forward layers. The output of the network is a predicted tag sequence in the IOB2 format. The aspect term that is to be predicted is outlined in the input sentence.}
  \label{fig:aspectmodel}
\end{figure}

We design a neural network based sequence tagger that reads in a sequence of words and predicts a sequence of corresponding IOB2 tags that encode the detected aspect terms.
Figure \ref{fig:aspectmodel} depicts the neural network component.

\subsubsection{Neural Network Sequence Tagger}
\label{sec:aspectterm}
The procedure to generate a tag sequence for a given word sequence can be described as follows:
First, the sequence of words is mapped to a sequence of word embedding vectors
$[w]_1^N=\{w_1,\ldots, w_N\}$, sentic vectors $[s]_1^N=\{s_1,\ldots, s_N\}$ and POS tag vectors $[p]_1^N=\{p_1,\ldots, p_N\}$ using the resources described in Section \ref{sec:features}.
We concatenate each word vector with its corresponding sentic vector and POS tag vector to receive the sequence:
$$[u]_1^N=\{u_1,\ldots, u_N\}= \{(w_1,s_1,p_1)^T, \ldots, (w_N,s_N,p_N)^T\} \text{ with } u_i \in \mathbb{R}^{100+5 + 45}.$$
The resulting sequence is passed to a bidirectional layer \cite{Schuster1997} of Gated Recurrent Units (GRU, \cite{cho2014learning}) that produces an output sequence of recurrent states:
$$[g]_1^N=\textsc{BiGRU}([u]_1^N)=\{g_1,\ldots, g_N\} \text{ with } g_i \in \mathbb{R}^{50},$$
using a combination of update and reset gates in each recurrent hidden unit.
Despite its simpler architecture and less demanding computations, the GRU is shown to be a competitive alternative to the well-known Long Short-Term Memory \cite{Chung2014}.
In practice, we implement the bidirectional GRU layer as two separate GRU layers.
One layer processes the input sequence in a forward direction (left-to-right) while the other processes it in reversed order (right-to-left).
The sequences of hidden states of each GRU layer are concatenated element wise in order to yield a single sequence of hidden states:
$$[g]_1^N= \{(\overrightarrow{g}_1,\overleftarrow{g}_1)^T, \ldots, (\overrightarrow{g}_N,\overleftarrow{g}_N)^T\} \text{ with } \overrightarrow{g}_i, \overleftarrow{g}_i \in \mathbb{R}^{25},$$
where $\overrightarrow{g}_i$ and $\overleftarrow{g}_i$ are the hidden states for the forward and backward GRU layer, respectively.
Each hidden state $g_i$ is passed to a regular feed-forward layer that produces a further hidden representation $h'_i \in \mathbb{R}^{50}$ for that state.
Lastly, a final layer in the network projects each $h'_i$ of the previous layer to a probability distribution $q_i$ over all possible output tags, namely I, O or B, using a softmax activation function:
$$[q]_1^N=\{q_1,\ldots q_N\} \text{ with } q_i \in \mathbb{R}^{3}.$$
For each word, we choose the tag with the highest probability as its predicted IOB2 tag.

Since the prediction of each tag can be interpreted as a classification, the network is trained to minimize the categorical cross-entropy between expected tag distribution $p_i$ and predicted tag distribution $q_i$ of each word $i$:
$$H(p_i,q_i) = -  \sum_{t \in \mathcal{T}} p_i(t) \log(q_i(t)),$$
where $\mathcal{T}=\{I,O, B\}$ is the set of IOB2 tags, $p_i(t) \in \{0,1\}$ is the expected probability of tag $t$ and $q(t)\in [0,1]$ the predicted probability.
The network's parameters are optimized using the stochastic optimization technique \emph{Adam} \cite{Kingma2014}.

For further processing, a predicted tag sequence can be decoded into aspect term annotations using the IOB2 scheme in reverse.
Note that we do not enforce the syntactic correctness of the predicted IOB2 scheme on a network-level.
It is possible that the network produces a tag sequence that is not correct in terms of the employed IOB2 scheme.
Thus, we post process each predicted tag sequence such that it constitutes a valid IOB2 tag sequence.
Specifically, we replace each $I$ tag that follows an $O$ tag with a $B$ in order to properly mark the beginning of an aspect term.

\subsection{Aspect-Specific Sentiment Extraction}
\label{sec:polarity}
The second step in our two-step architecture for aspect-based sentiment extraction is the prediction of a polarity label given a previously detected aspect term.
We address this aspect-specific sentiment extraction using a recurrent neural network that is, in parts, very similar to the architecture for aspect term extraction in Section \ref{sec:aspectterm}.

\begin{figure}[t]
  \centering
  \includegraphics[width=0.75\columnwidth]{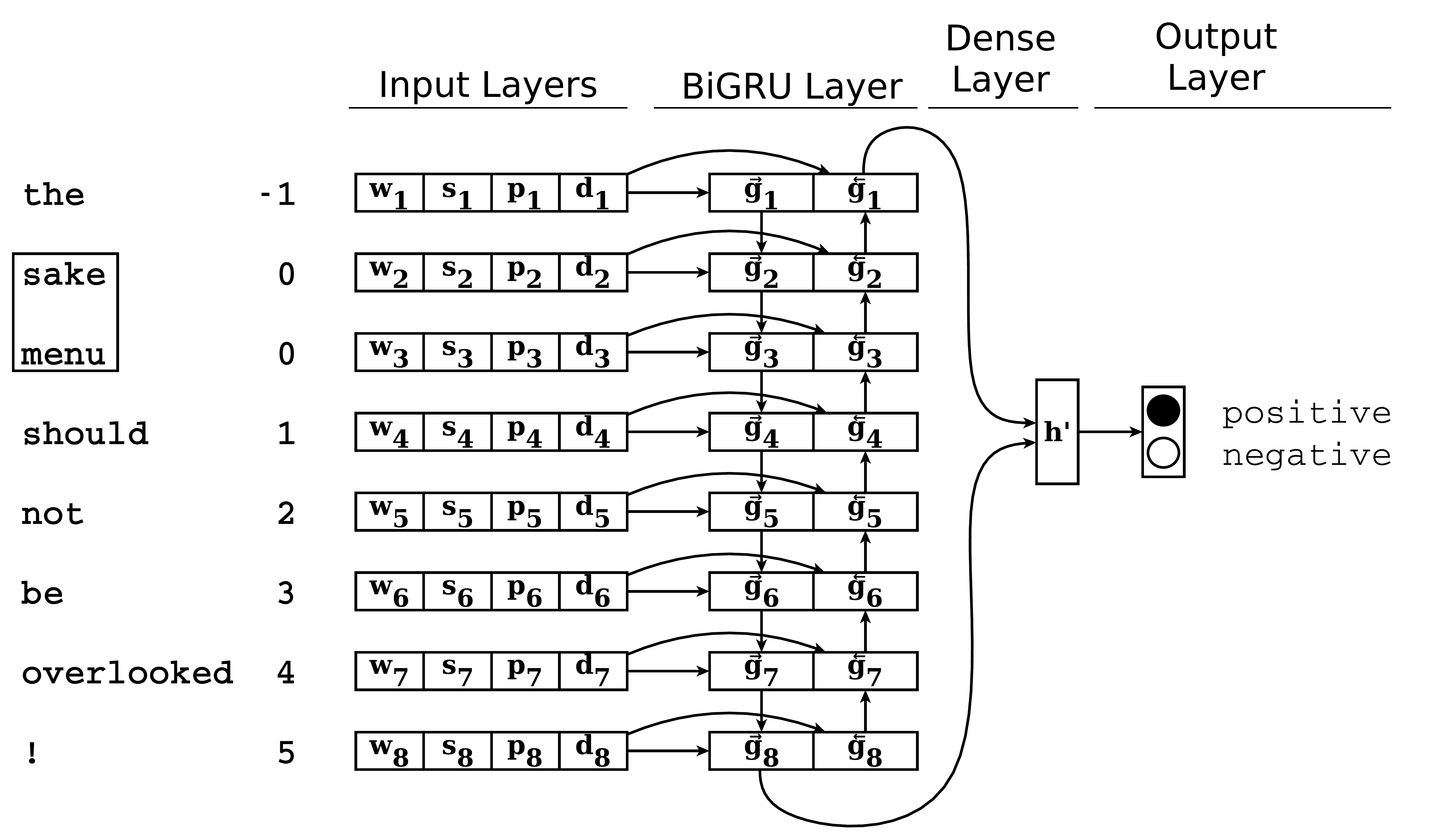}
  \caption{The aspect-specific sentiment extraction component. The network processes the input sentence as a sequence of word vectors $w_i$, sentic vectors $s_i$, POS tags $p_i$ and distance embeddings $d_i$ using a bidirectional GRU layer and regular feed-forward layers. The output of the network is a single predicted polarity label for the aspect term of interest. The aspect term for which a polarity label is to be predicted is outlined in the input sentence.}
  \label{fig:polaritymodel}
\end{figure}

In order to predict a polarity label for a \emph{specific} aspect term in a sentence, we need to mark the aspect term in question.
For this, we apply a similar technique as has been done for relation extraction \cite{Zeng2014} and Semantic Role Labeling \cite{collobert2011natural}.
We tag each word in the input sentence with its relative distance to the aspect term, as follows:
\begin{center}
\small
\begin{tabular}{WWWWWW}
Great & \textbf{service} & , & great & \textit{food} & . \\
-1 & 0 & 1 & 2 & 3 & 4
\end{tabular}
\end{center}
where the bold word ``\textbf{service}'' is the aspect term for which we want to extract the polarity.
The italic word ``\textit{food}'' marks another aspect term.
The relative distance to the selected aspect term is shown below each word.
This sequence of relative distances implicitly encodes the position of the aspect term in question in the sentence.
In theory, this strategy permits to incorporate long range information in the prediction process in contrast to cutting a fixed-sized (and usually small) window of words around the aspect term in the sentence.
In practice, we do not use the raw distance values directly but represent them as 10 dimensional distance embedding vectors similar as in \cite{Xiang2015,Zeng2014,Sun2015} and treat them as learnable parameters in our network.
We further denote the sequence of distance embedding vectors for a sentence of $N$ words as:
$$[d]_1^N= \{d_1, \ldots, d_N\} \text{ with } d_i \in \mathbb{R}^{10}.$$
Figure \ref{fig:polaritymodel} depicts the neural network component.

\subsubsection{Neural Network Polarity Extraction}
The procedure for predicting a polarity label for an aspect term can be described as follows:
Assume we have a sentence and an already extracted aspect term.
We concatenate each word vector with its corresponding sentic vector, its POS tag vector and distance vector to receive the sequence:
$$[u]_1^N= \{(w_1,s_1,p_1, d_1)^T, \ldots, (w_N,s_N,p_N,d_N)^T\} \text{ with } u_i \in \mathbb{R}^{100+5+45+10}.$$
The resulting sequence is passed to a bidirectional GRU layer that produces an output sequence of recurrent states:
$$[g]_1^N=\textsc{BiGRU}([u]_1^N)= \{(\overrightarrow{g}_1,\overleftarrow{g}_1)^T, \ldots, (\overrightarrow{g}_N,\overleftarrow{g}_N)^T\} \text{ with } \overrightarrow{g}_i, \overleftarrow{g}_i \in \mathbb{R}^{25}.$$
We take the final hidden state $\overrightarrow{g}_N$ of the forward GRU and the final hidden state $\overleftarrow{g}_1$ of the backward GRU\footnote{Since this GRU processes the sequence in a reversed direction, the final hidden state is the hidden state for the first word.} and concatenate them to receive a fixed sized representation $h=(\overrightarrow{g}_N,\overleftarrow{g}_1)^T \in \mathbb{R}^{50}$ of the aspect term in the whole input sentence.
Next, the network passes the hidden representation $h$ of the aspect term through a densely connected feed-forward layer producing another hidden representation $h' \in \mathbb{R}^{50}$.
As a last step, a final densely connected layer with a softmax activation function projects $h'$ to a 2-dimensional vector $q\in \mathbb{R}^{2}$ representing a probability distribution over the two polarity labels \texttt{positive} and \texttt{negative}.
We consider the label with the highest estimated probability to be the predicted polarity label for the given aspect term.

Again, we train the network to minimize the categorical cross-entropy between expected polarity label distribution $p$ and predicted polarity label distribution $q$ of each aspect term:
$$H(p,q) = -  \sum_{l \in \mathcal{L}} p(l) \log(q(l)),$$
where $\mathcal{L}=\{\texttt{positive},\texttt{negative}\}$ is the set of polarity labels and $p(l)$ and $q(l)$ the expected and predicted probability, respectively, for label $l$.
As before, we apply the \emph{Adam} technique to update network parameters.

\section{Experiments and Evaluation}
\label{sec:experiments}
In order to see the performance of the overall system and the impact of the individual features, we perform an evaluation on the provided training data for the aspect-based sentiment analysis task.
Based on that we select a final model configuration that is used in the actual challenge evaluation on additional test data.

\subsubsection{Evaluation on Training Data}
All experiments on the training data are performed as a 5-fold cross-validation.
We evaluate the two steps of our approach separately to better see the individual performances of the two components.

Since we do not have access to official evaluation scripts, we evaluate aspect term extraction 
using Precision, Recall and F$_1$ score.
We only take explicitly mentioned aspect terms into account\footnote{We exclude annotations with \texttt{aspect=}``\texttt{NULL}''.} that have a polarity label of either \texttt{positive} or \texttt{negative}.
Identical annotations i.e. annotations that target the same aspect term (in terms of character offsets) with the same polarity, are considered as one.
Table \ref{tab:aspect} shows the results for aspect term extraction for different feature combinations.
\begin{table}[t]
\centering
\begin{tabular}{lC{2cm}C{2cm}C{2cm}}
\toprule
Features & F$_1$ & Precision & Recall  \\
\midrule
\texttt{WE+POS}                          & \textbf{0.684} & 0.659         & \textbf{0.710}             \\ 
\texttt{WE+POS+Sentics}            & 0.679 & \textbf{0.663}    & 0.697      \\ 
\texttt{WE-Retro+POS}           & 0.678 & 0.651         & 0.708             \\ 
\texttt{WE-Retro+POS+Sentics} & 0.679 & 0.655         & 0.706             \\ 
\bottomrule
\end{tabular}
\vspace{0.5cm}
\caption{Results of 5-fold cross-validation for aspect term extraction using different feature combinations.}
\label{tab:aspect}
\end{table}
Here, \texttt{WE} denotes the usage of amazon review word embeddings, \texttt{WE-Retro} denotes the retrofitted embeddings, \texttt{POS} specifies additional POS tag features and \texttt{Sentics} indicates the usage of sentic vectors.

Comparing the models in Table \ref{tab:aspect}, we can see that using the retrofitted embeddings seems to downgrade the performance of our system.
Also, employing the sentic vectors for aspect term extraction degrades the networks performance.
This is not completely unexpected, though, since the sentic vectors mainly encode sentiment information and aspect term extraction on its own is rather decoupled from the actual sentiment extraction.
A more positive effect would be expected for the second step in our system, the prediction of polarity labels.

To evaluate the aspect-specific sentiment extraction, we extract polarity labels for all aspect terms of the ground truth annotations.
By separating aspect term extraction and sentiment extraction, we can better evaluate the sentiment extraction in isolation.
Again, we only consider unique aspect terms that are either labeled with a \texttt{positive} or \texttt{negative} polarity.
We report the performance of our sentiment extraction in terms of the accuracy of the system for different feature combinations.
Table \ref{tab:polarity} shows the results for the 5-fold cross-validation on the training data.
\texttt{WE}, \texttt{WE-Retro}, \texttt{POS} and \texttt{Sentics} are defined as before, while \texttt{Dist} denotes the obligatory distance embedding features.
\newfloatcommand{capbtabbox}{table}[][0.48\textwidth]
\begin{table}[t]
  \begin{floatrow}
    \capbtabbox{%
    \begin{tabular}{lC{1.5cm}}
    \toprule
    Features & Accuracy \\
    \midrule
    \texttt{WE+POS+Dist}                          & 0.776       \\
    \texttt{WE+POS+Dist+Sentics}            & \textbf{0.811} \\ 
    \texttt{WE-Retro+POS+Dist}            	  & 0.776       \\ 
    \texttt{WE-Retro+POS+Dist+Sentics}  & 0.809       \\ 
    \bottomrule
    \end{tabular}
  }{%
    \vspace{0.2cm}%
    \caption{Results of 5-fold cross-validation for aspect-specific sentiment extraction using different feature combinations.}%
    \label{tab:polarity}%
  }
  \ffigbox{%
    \includegraphics[width=0.48\textwidth,trim=1cm 0.8cm 1cm 1cm]{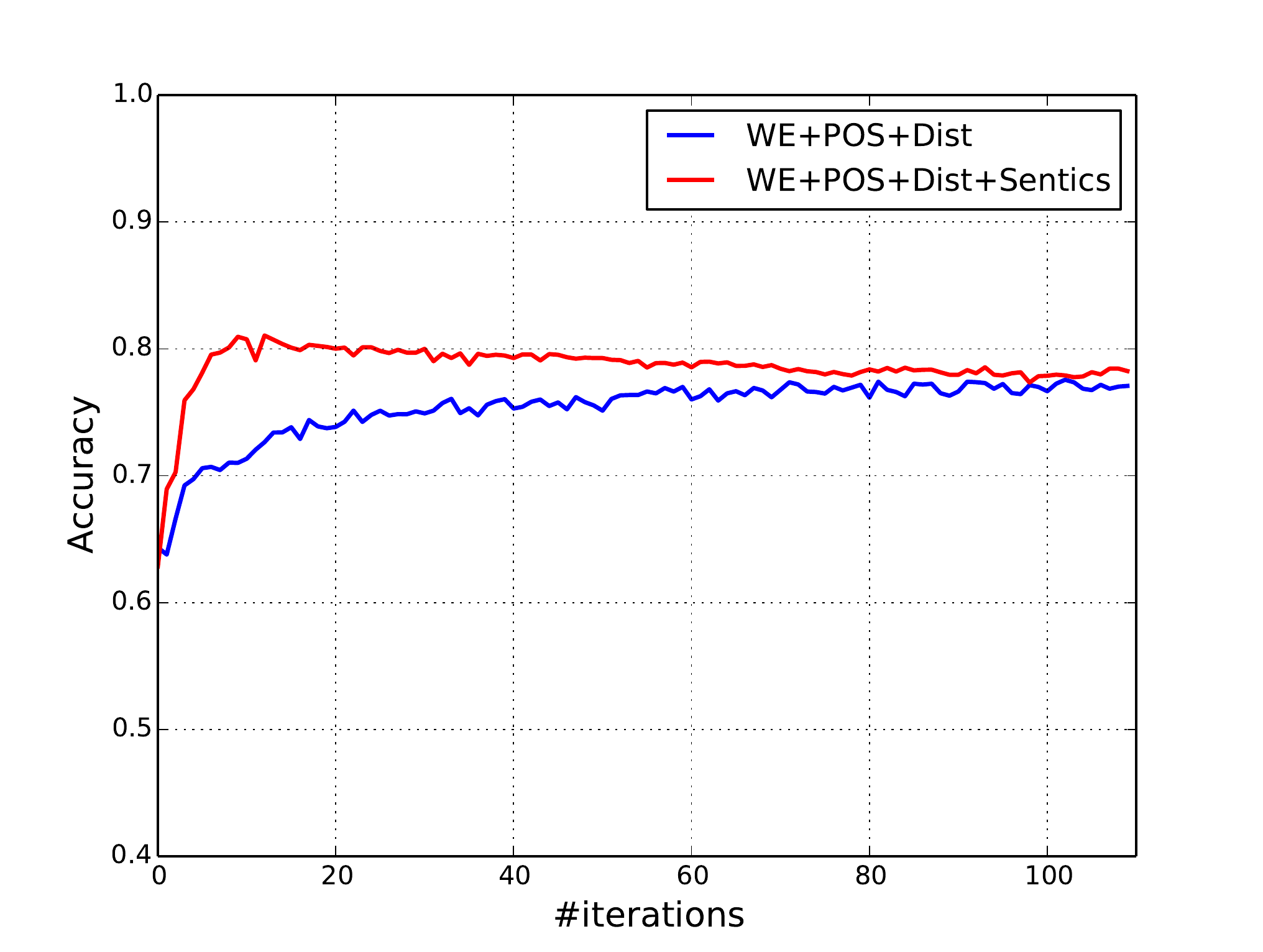}
  }{%
    \caption{Visualization of the performance gain of using sentic vectors with respect to the number of iterations over the training data. By using additional sentic vectors we achieve better results with less training needed.}%
    \label{fig:iterations}%
  }
  \end{floatrow}
\end{table}

While the retrofitted embeddings do not contribute positively to the performance for sentiment extraction either, a notable gain is achieved using the sentic vectors in our component for aspect-specific sentiment extraction.
Here, we observe a gain of 3.5\% points accuracy compared to using only word embeddings, distance embeddings and POS tags.
Apart from that, the usage of sentic vectors drastically reduces the training time needed to achieve these results.
The best results for the \texttt{WE+POS+Dist} and \texttt{WE-Retro+POS+Dist} model were achieved with 102 iterations over the training portion of the data, while the \texttt{WE+POS+Dist+Sentic} and \texttt{WE-Retro+POS+Dist+Sentic} model reached their best performances for only 12 and 9 iterations, respectively. 
See Figure \ref{fig:iterations} for a visualization of the system's accuracy with respect to the employed features and the iteration over the training data.

\subsubsection{Evaluation on Test Data}
Apart from our custom evaluation, each participating system is evaluated on a separate test set of customer reviews as part of the Sentiment Analysis Challenge.
While the annotated training data covers the domains \emph{laptops} and \emph{restaurants}, the data for the test set is obtained from the domains \emph{restaurants} and \emph{hotels} in order to test the systems on a previously unseen review domain.
For comparability, the predicted results for each system are evaluated by the organizers.
Aspect term extraction is evaluated with precision, recall and F$_1$ score regarding exact matches.
Polarity extraction is evaluated with the accuracy of the predicted polarity label with respect to the subset of correctly extracted aspect terms from the previous step.

For this evaluation, we train final models for our two architectural components using knowledge gained from our preliminary results on the training data.
The aspect term extract model \texttt{WE+POS} is trained on all training samples for 5 epochs and the polarity extraction model \texttt{WE+POS+Dist+Sentics} for 10 epochs.
The official evaluation on the test data shows an F$_1$ score of 0.433 with a precision of 0.415 and a recall of 0.452 for the aspect term extraction in separation.
The extraction of aspect-specific polarity labels for correctly identified aspect terms results in an accuracy of 0.874.
With these results, the proposed system achieves the highest scores of the 2016th ESWC fine-grained sentiment analysis challenge.

\section{Conclusion}
With this work we propose a two-step approach for aspect-based sentiment analysis.
We decouple the extraction of aspects and sentiment labels in order to obtain a flexibly applicable system.
By using a recurrent neural network, we present a novel neural network based approach to tackle aspect extraction as a sequence labeling task.
Furthermore, we present a novel way to address aspect-specific sentiment extraction using a recurrent neural network architecture with distance embedding features.
This model is able to extract sentiments expressed towards a specific aspect that is mentioned in the text and is thus able to detect multiple opinions in a single sentence.

Both components of our overall sentiment analysis system incorporate additional semantic knowledge by using pretrained word vectors that are \emph{retrofitted} to a semantic lexicon as well as semantic and sentiment-related features obtained from SenticNet.
Although our first experiments could not show a benefit in using the retrofitted embeddings, the sentics obtained from SenticNet proved to be a valuable feature for extracting aspect-based polarity labels that increased accuracy and shortened training time considerably.

For this work, we could only incorporate single-word concepts from SenticNet as additional features.
For the future, we plan to modify our architecture to permit incorporation of all concepts from SenticNet, thus moving the system to concept-level sentiment analysis even further.

\section*{Acknowledgements}
This work was supported by the Cluster of Excellence Cognitive Interaction Technology 'CITEC' (EXC 277) at Bielefeld University, which is funded by the German Research Foundation (DFG).

\bibliographystyle{splncs03}
\bibliography{bibliography.bib}

\begin{thebibliography}{10}
\providecommand{\url}[1]{\texttt{#1}}
\providecommand{\urlprefix}{URL }

\bibitem{AprosioCDR15}
Aprosio, A.P., Corcoglioniti, F., Dragoni, M., Rospocher, M.: Supervised
  opinion frames detection with {RAID}. In: Gandon, F., Cabrio, E., Stankovic,
  M., Zimmermann, A. (eds.) Semantic Web Evaluation Challenges - Second
  SemWebEval Challenge at {ESWC} 2015, Portoro{\v{z}}, Slovenia. Communications
  in Computer and Information Science, vol. 548, pp. 251--263. Springer (2015)

\bibitem{Cambria2014sentic}
Cambria, E., Olsher, D., Rajagopal, D.: {SenticNet 3: a common and common-sense
  knowledge base for cognition-driven sentiment analysis}. In: Proceedings of
  the Twenty-Eighth AAAI Conference on Artificial Intelligence. pp. 1515--1521
  (2014)

\bibitem{cho2014learning}
Cho, K., Van~Merri{\"{e}}nboer, B., G{\"{u}}l{\c c}ehre, {\c C}., Bahdanau, D.,
  Bougares, F., Schwenk, H., Bengio, Y.: Learning phrase representations using
  rnn encoder--decoder for statistical machine translation. In: Proceedings of
  the 2014 Conference on Empirical Methods in Natural Language Processing
  (EMNLP). pp. 1724--1734. Association for Computational Linguistics, Doha,
  Qatar (Oct 2014)

\bibitem{ChungWT14}
Chung, J.K., Wu, C., Tsai, R.T.: Polarity detection of online reviews using
  sentiment concepts: {NCU} {IISR} team at {ESWC-14} challenge on concept-level
  sentiment analysis. In: Presutti, V., Stankovic, M., Cambria, E., Cantador,
  I., Iorio, A.D., Noia, T.D., Lange, C., Recupero, D.R., Tordai, A. (eds.)
  Semantic Web Evaluation Challenge - SemWebEval 2014 at {ESWC} 2014,
  Anissaras, Crete, Greece. Communications in Computer and Information Science,
  vol. 475, pp. 53--58. Springer (2014)

\bibitem{Chung2014}
Chung, J., G{\"{u}}l{\c{c}}ehre, {\c{C}}., Cho, K., Bengio, Y.: Empirical
  evaluation of gated recurrent neural networks on sequence modeling. In: NIPS
  Deep Learning Workshop (2014)

\bibitem{collobert2011natural}
Collobert, R., Weston, J., Bottou, L., Karlen, M., Kavukcuoglu, K., Kuksa, P.:
  Natural language processing (almost) from scratch. Journal of Machine
  Learning Research  12,  2493--2537 (2011)

\bibitem{DragoniTP14}
Dragoni, M., Tettamanzi, A.G.B., da~Costa~Pereira, C.: A fuzzy system for
  concept-level sentiment analysis. In: Presutti, V., Stankovic, M., Cambria,
  E., Cantador, I., Iorio, A.D., Noia, T.D., Lange, C., Recupero, D.R., Tordai,
  A. (eds.) Semantic Web Evaluation Challenge - SemWebEval 2014 at {ESWC} 2014,
  Anissaras, Crete, Greece. Communications in Computer and Information Science,
  vol. 475, pp. 21--27. Springer (2014)

\bibitem{Faruqui2015}
Faruqui, M., Dodge, J., Jauhar, S.K., Dyer, C., Hovy, E., Smith, N.A.:
  {Retrofitting Word Vectors to Semantic Lexicons}. Proceedings of Human
  Language Technologies: The 2015 Annual Conference of the North American
  Chapter of the ACL pp. 1606--1615 (2015)

\bibitem{fellbaum2005}
Fellbaum, C.: Wordnet and wordnets. In: Brown, K. (ed.) Encyclopedia of
  Language and Linguistics. pp. 665--670. Elsevier, Oxford (2005)

\bibitem{Hu2004}
Hu, M., Liu, B.: Mining and summarizing customer reviews. In: Proceedings of
  the Tenth ACM SIGKDD International Conference on Knowledge Discovery and Data
  Mining (KDD '04). pp. 168--177. ACM, New York, NY, USA (2004)

\bibitem{Jakob2010}
Jakob, N., Gurevych, I.: {Extracting opinion targets in a single-and
  cross-domain setting with conditional random fields}. In: Proceedings of the
  Conference on Empirical Methods in Natural Language Processing. pp.
  1035--1045 (October 2010)

\bibitem{Kingma2014}
Kingma, D., Ba, J.: {Adam: A Method for Stochastic Optimization}. International
  Conference on Learning Representations  (2015)

\bibitem{Klinger2013a}
Klinger, R., Cimiano, P.: Bi-directional inter-dependencies of subjective
  expressions and targets and their value for a joint model. In: Proceedings of
  the 51st Annual Meeting of the Association for Computational Linguistics
  (ACL), Volume 2: Short Papers. pp. 848--854 (August 2013)

\bibitem{Klinger2013b}
Klinger, R., Cimiano, P.: Joint and pipeline probabilistic models for
  fine-grained sentiment analysis: Extracting aspects, subjective phrases and
  their relations. In: Proceedings of the 13th {IEEE} International Conference
  on Data Mining Workshops (ICDM). pp. 937--944 (December 2013)

\bibitem{lakkaraju2014aspect}
Lakkaraju, H., Socher, R., Manning, C.: {Aspect Specific Sentiment Analysis
  using Hierarchical Deep Learning}. Proceedings of the NIPS Workshop on Deep
  Learning and Representation Learning  (2014)

\bibitem{Le2014}
Le, Q., Mikolov, T.: {Distributed Representations of Sentences and Documents}.
  ICML  32,  1188--1196 (2014)

\bibitem{Manning2014}
Manning, C., Surdeanu, M., Bauer, J., Finkel, J., Bethard, S., McClosky, D.:
  {The Stanford CoreNLP Natural Language Processing Toolkit}. In: Proceedings
  of 52nd Annual Meeting of the Association for Computational Linguistics:
  System Demonstrations. pp. 55--60 (2014)

\bibitem{McAPanLes15}
McAuley, J., Pandey, R., Leskovec, J.: Inferring networks of substitutable and
  complementary products. In: Proceedings of the 21th ACM SIGKDD International
  Conference on Knowledge Discovery and Data Mining (KDD '15). pp. 785--794.
  ACM, New York, NY, USA (2015)

\bibitem{McATarShiHen15}
McAuley, J., Targett, C., Shi, Q., van~den Hengel, A.: Image-based
  recommendations on styles and substitutes. In: Proceedings of the 38th
  International ACM SIGIR Conference on Research and Development in Information
  Retrieval. pp. 43--52. ACM (2015)

\bibitem{Mikolov2013Efficient}
Mikolov, T., Corrado, G., Chen, K., Dean, J.: {Efficient Estimation of Word
  Representations in Vector Space}. In: Proceedings of the International
  Conference on Learning Representations (2013)

\bibitem{Mikolov2013Distributed}
Mikolov, T., Sutskever, I., Chen, K., Corrado, G.S., Dean, J.: Distributed
  representations of words and phrases and their compositionality. In: Advances
  in Neural Information Processing Systems. pp. 3111--3119 (2013)

\bibitem{pennington2014glove}
Pennington, J., Socher, R., Manning, C.: {Glove: Global Vectors for Word
  Representation}. In: Proceedings of the 2014 Conference on Empirical Methods
  in Natural Language Processing (EMNLP). pp. 1532--1543. Association for
  Computational Linguistics, Doha, Qatar (2014)

\bibitem{Pontiki2015}
Pontiki, M., Galanis, D., Papageorgiou, H., Manandhar, S., Androutsopoulos, I.:
  Semeval-2015 task 12: Aspect based sentiment analysis. In: Proceedings of the
  9th International Workshop on Semantic Evaluation. pp. 486--495. Association
  for Computational Linguistics, Denver, Colorado (June 2015)

\bibitem{agerri2015elixa}
San~Vicente, I.n., Saralegi, X., Agerri, R.: Elixa: A modular and flexible
  {ABSA} platform. In: Proceedings of the 9th International Workshop on
  Semantic Evaluation. pp. 748--752. Association for Computational Linguistics,
  Denver, Colorado (June 2015)

\bibitem{santos2014learning}
dos Santos, C., Zadrozny, B.: Learning character-level representations for
  part-of-speech tagging. In: Proceedings of the 31st International Conference
  on Machine Learning. pp. 1818--1826 (2014)

\bibitem{Xiang2015}
dos Santos, C.N., Xiang, B., Zhou, B.: Classifying relations by ranking with
  convolutional neural networks. In: Proceedings of the 53rd Annual Meeting of
  the Association for Computational Linguistics and the 7th International Joint
  Conference on Natural Language Processing. vol.~1, pp. 626--634 (2015)

\bibitem{SchoutenF15}
Schouten, K., Frasincar, F.: The benefit of concept-based features for
  sentiment analysis. In: Gandon, F., Cabrio, E., Stankovic, M., Zimmermann, A.
  (eds.) Semantic Web Evaluation Challenges - Second SemWebEval Challenge at
  {ESWC} 2015, Portoro{\v{z}}, Slovenia. Communications in Computer and
  Information Science, vol. 548, pp. 223--233. Springer (2015)

\bibitem{Schuster1997}
Schuster, M., Paliwal, K.K.: {Bidirectional recurrent neural networks}. IEEE
  Transactions on Signal Processing  45(11),  2673--2681 (1997)

\bibitem{Sun2015}
Sun, Y., Lin, L., Tang, D., Yang, N., Ji, Z., Wang, X.: Modeling mention,
  context and entity with neural networks for entity disambiguation. In:
  Proceedings of the 24th International Conference on Artificial Intelligence
  (IJCAI). pp. 1333--1339. AAAI Press (2015)

\bibitem{Titov2008}
Titov, I., Mcdonald, R.: {A Joint Model of Text and Aspect Ratings for
  Sentiment Summarization}. In: Proceedings of Annual Meeting of the
  Association for Computational Linguistics (ACL). pp. 308--316 (2008)

\bibitem{tksveenstra99eacl}
Tjong Kim~Sang, E.F., Veenstra, J.: Representing text chunks. In: Proceedings
  of European Chapter of the ACL (EACL). pp. 173--179. Bergen, Norway (1999)

\bibitem{Toh2014}
Toh, Z., Wang, W.: {DLIREC: Aspect Term Extraction and Term Polarity
  Classification System}. In: Proceedings of the 8th International Workshop on
  Semantic Evaluation. pp. 235--240 (2014)

\bibitem{Zeng2014}
Zeng, D., Liu, K., Lai, S., Zhou, G., Zhao, J.: {Relation Classification via
  Convolutional Deep Neural Network}. In: Proceedings of the 25th International
  Conference on Computational Linguistics (COLING). pp. 2335--2344 (2014)

\end{thebibliography}

\end{document}